\documentclass[letterpaper, 10 pt, conference]{ieeeconf}  

\IEEEoverridecommandlockouts                              

\usepackage{graphicx}
\usepackage{booktabs}
\usepackage{makecell} 
\usepackage{hyperref}

\usepackage[dvipsnames]{xcolor}
\title{\LARGE \bf
Leveraging LLMs with Iterative Loop Structure for Enhanced Social Intelligence in Video Question Answering
}

\author{Erika Mori$^{1,2}$, Yue Qiu$^{1}$, Hirokatsu Kataoka$^{1}$ and Yoshimitsu Aoki$^{2}$ \\
$^{1}$National Institute of Advanced Industrial Scienece and Technology (AIST), $^{2}$Keio University
}
\begin{document}

\maketitle
\thispagestyle{empty}
\pagestyle{empty}

\begin{abstract}

Social intelligence, the ability to interpret emotions, intentions, and behaviors, is essential for effective communication and adaptive responses. As robots and AI systems become more prevalent in caregiving, healthcare, and education, the demand for AI that can interact naturally with humans grows. However, creating AI that seamlessly integrates multiple modalities, such as vision and speech, remains a challenge. Current video-based methods for social intelligence rely on general video recognition or emotion recognition techniques, often overlook the unique elements inherent in human interactions. To address this, we propose the Looped Video Debating (LVD) framework, which integrates Large Language Models (LLMs) with visual information, such as facial expressions and body movements, to enhance the transparency and reliability of question-answering tasks involving human interaction videos. Our results on the Social-IQ 2.0 benchmark show that LVD achieves state-of-the-art performance without fine-tuning. Furthermore, supplementary human annotations on existing datasets provide insights into the model's accuracy, guiding future improvements in AI-driven social intelligence. The proposed dataset can be accessed via the following \href{https://github.com/edrkr96/Social-IQ-2.0-Sub}{link (https://github.com/edrkr96/Social-IQ-2.0-Sub)}.


\end{abstract}

\section{INTRODUCTION}


Social intelligence is a distinct human ability encompassing a broad range of skills necessary for successful social interactions and appropriate behavior across various contexts. It allows individuals to recognize and interpret others’ emotions, intentions, and behavioral patterns, enabling nuanced and adaptive responses in everyday situations. For robotics to effectively integrate into human society and interact in daily life, understanding social intelligence will be crucial. While social intelligence has been extensively studied in fields such as sociology and psychology, where its role in human cognition and behavior is well-documented \cite{si-1920, si-1928, si-1937, si-1965, si-1973, si-1999, si-2007}, it remains underexplored in the domains of AI and robotics.


The recent integration of robots into caregiving, healthcare, and education, coupled with the widespread adoption of conversational AI tools like ChatGPT \cite{chatgpt}, has created a growing demand for robots and AI systems capable of natural human-like interactions. To fulfill this demand, these systems need to be equipped with social intelligence to understand human emotions, interpret intentions, and respond appropriately. However, the development of AI that effectively utilizes multiple modalities, such as vision and speech, for seamless human communication remains a significant challenge \cite{singh2023artificial}.

Several key challenges exist in developing socially intelligent robots and AI tools. While conversational AI models like ChatGPT, trained on vast datasets, have demonstrated some degree of social intelligence by generating responses that account for cultural contexts, manners, and user intentions, there are few examples where large language models (LLMs) are explicitly used to understand multimodal human interactions. Most existing approaches to understanding human interactions in videos \cite{just-ask-plus, multi} are extensions of general video comprehension or emotion recognition techniques and do not explicitly extract diverse information critical for social understanding, such as dialogue content, facial expressions, voice tone, and scene context. Furthermore, many of these approaches \cite{just-ask-plus, multi, desiq} are end-to-end models, which often lack transparency and reliability, particularly for complex tasks like question-answering based on human interaction videos.

\begin{figure}[t]
  \centering
  \includegraphics[width=\linewidth]{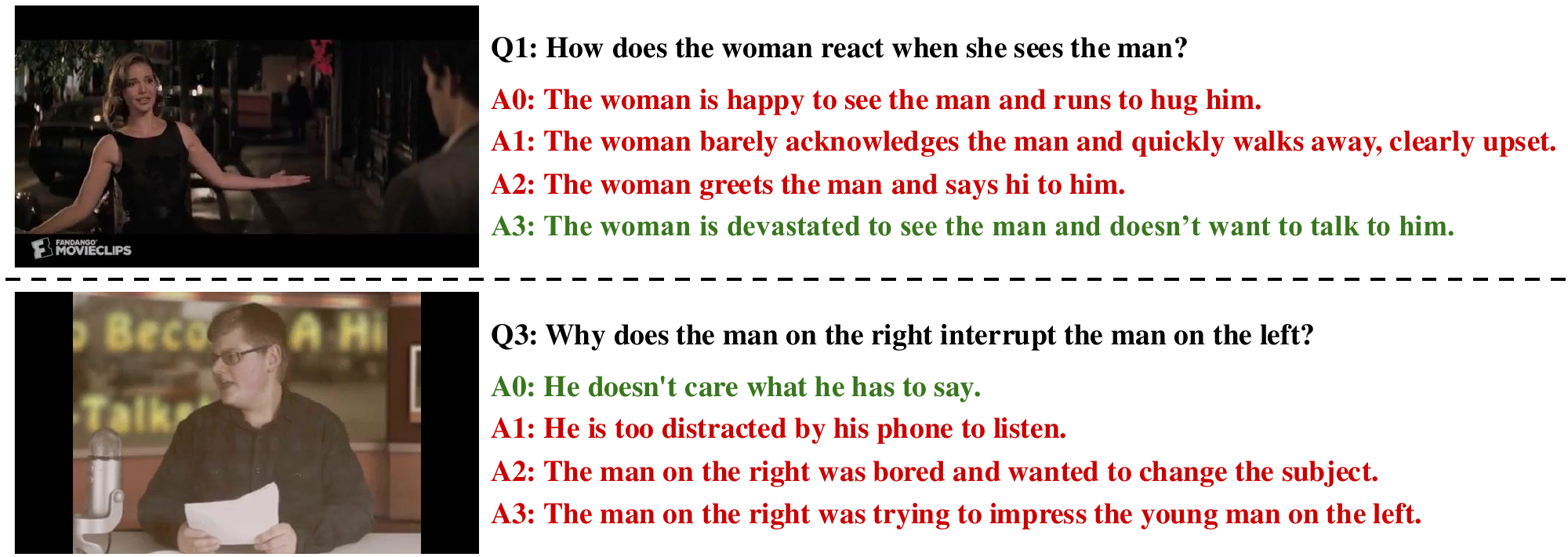} 
  \caption{Two samples from the Social-IQ 2.0 dataset \cite{siq2}. Each video, depicting human interactions, is accompanied by approximately six questions that require advanced reasoning. For each question, one correct answer (\textcolor{OliveGreen}{green}) and three incorrect answers (\textcolor{red}{red}) are provided.}
  \label{fig:Social-IQ}
\end{figure}

To address these challenges, we propose a novel framework called Looped Video Debating (LVD), which leverages the social intelligence capabilities of LLMs while explicitly incorporating detailed visual information, such as facial expressions and body movements. LVD is designed as a looped framework for question-answering tasks related to human interaction videos, combining an LLM with a visual question answering (VQA) model. This design improves transparency and reliability by enabling the model to produce not only answers but also rationale and additional information required.

Our experimental results on the Social-IQ 2.0 benchmark \cite{siq2} show that LVD achieves state-of-the-art performance without fine-tuning. Furthermore, to assess the accuracy of the generated rationale and additional information, we conduct supplementary annotations on existing datasets and compare the performance between humans and LLMs in detail. This analysis offers valuable insights for future enhancements of the model.


\section{Related Work}
\subsection{Social Intelligence Recognition in AI}
Social intelligence is the ability to understand others' intentions and the atmosphere of a situation through verbal and non-verbal communication and to behave appropriately in social interactions. This ability is essential for natural interactions between robots and humans, and it has become a focus of active research in the fields of robotics and AI.

In robotics, social intelligence is often evaluated through simulations. For example, Lee et al. \cite{lee2023developing} measure a robot's social intelligence using counseling simulations. They assess how well the robot-generated responses match human responses in categories such as speech, action, facial expression, and emotion to evaluate how accurately the robot can replicate non-verbal empathy styles. Similarly, Zhou et al. \cite{zhou2023sotopia} evaluate the interactions of artificial agents in various social scenarios from multiple perspectives, such as goal achievement and relationship maintenance. Their study defines agent actions in each turn, including facial expressions and movements, enabling the evaluation of non-verbal information. However, these simulations are limited to predefined textual information and do not use images or videos, lacking detailed visual information.

In the field of computer vision, social intelligence is often evaluated through question answering (QA) tasks involving images or videos, as seen in benchmarks like VCR \cite{vcr}, MovieQA \cite{movieqa}, and TVQA \cite{tvqa}. VCR is limited to image-based inference, while MovieQA and TVQA include questions about people or actions in videos, typically beginning with ``what'' or ``who''. In contrast, Social-IQ \cite{siq} and its updated version, Social-IQ 2.0 \cite{siq2}, offer more challenging benchmarks by incorporating diverse modalities, such as videos, images, dialogues and audio, and presenting advanced questions that explore the causes behind specific emotions and the intentions behind behaviors. We use Social-IQ 2.0 as our evaluation benchmark due to its richness of modalities and the complexity of its questions. However, to evaluate aspects not covered in the existing Social-IQ 2.0 dataset, such as answer validity and rationale, we added additional annotations to a subset of the data.

\vskip\baselineskip


\begin{figure*}[t]
  \centering
  \includegraphics[width=\linewidth]{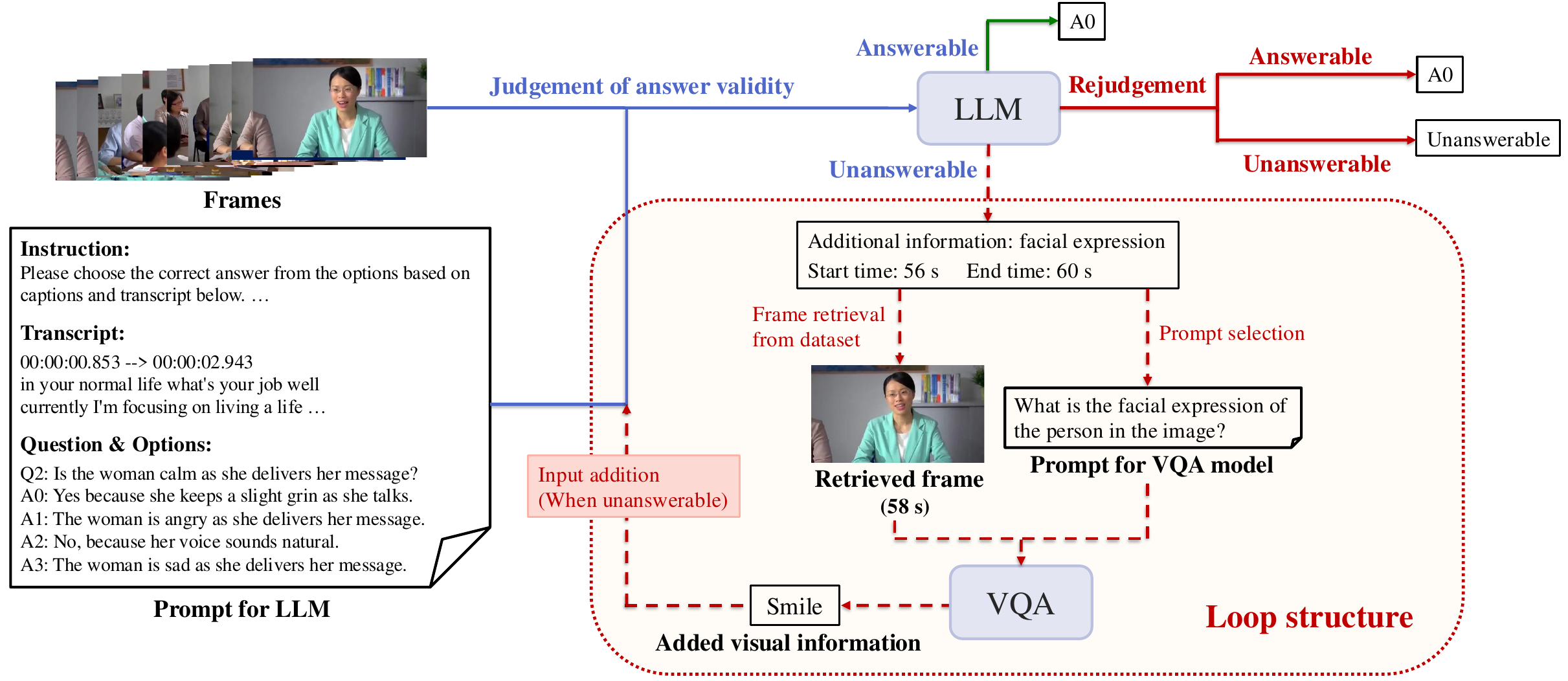}
  \caption{
  {\bf Proposed method (LVD).}
    In this method, the model first determines whether the question is answerable based on 10 images (or captions, in the case of GPT-4 and Llama) and the dialogue information (\textcolor{blue}{blue}). If the question is considered answerable, the option inferred to be correct is output (the \textcolor{OliveGreen}{green} arrow). If the question is deemed unanswerable, a loop structure is employed to obtain additional information (\textcolor{red}{red} dashed arrows). This additional information is then added to the original input, and the QA process is repeated (\textcolor{red}{red} solid arrows).
    }
  \label{fig:approach}
\end{figure*}

\subsection{Social Intelligence Recognition Method}  
Just Ask Plus is state-of-the-art method in Social-IQ 2.0 benchmark, one of the most widely used social intelligence datasets. This method focuses on acquiring multimodal representations and selecting feature extractors. It relies on choosing the generated option feature most similar to the question feature, making it challenging to answer atypical questions or those that depend heavily on video context. The most accurate method with training is the Multi-Modal Temporal Correlated Network with Emotional Social Cues (MMTC-ESC) \cite{multi}. This approach surpasses existing methods by utilizing multimodal inputs and contrastive learning with emotional features of question-option pairs. However, since it concentrates exclusively on emotions, it often fails to select the correct answer when the type of emotion alone is insufficient to justify the answer. Analysis in \cite{multi} indicates that there is only about a 25\% chance that the correct option's emotion is not present in the other three options. 

With the recent remarkable improvements, LLMs have developed the ability to recognize human interactions within text. Current research has assessed their social intelligence using text-based QA \cite{choi2023llms} and simulations with artificial agents \cite{zhou2023sotopia}. However, these studies are limited to textual modalities, treating non-verbal cues, such as facial expressions and gestures, as simple text descriptions.

At the same time, the application of LLMs to video-related tasks has been expanding \cite{liu2024tempcompass, pan2023retrieving, huang2024free, lin2023videodirectorgpt}. Despite this growth, recent studies show that video-based LLMs face significant challenges, including the integration of multiple modalities and the comprehension of long-duration videos \cite{video-chatgpt, video-llama, videogpt+, videollama2}. These limitations make them less effective for advanced content understanding tasks like Social-IQ 2.0 dataset.

To address these challenges, this study explores the potential for improving accuracy in an end-to-end framework for the Social-IQ 2.0 dataset by integrating an LLM with a VQA model.

\vskip\baselineskip

\section{Approach}
\label{sec:approach}
This study aims to evaluate social intelligence of LLMs by conducting QA using images and dialogue information from the Social-IQ 2.0 dataset under various settings. Additionally, to improve QA accuracy without additional training of LLM, we propose a framework called Looped Video Debating (LVD), which combines LLM with a VQA model. The overview of the proposed method is shown in Figure \ref{fig:approach}. As a novel setting, we included an ``unanswerable'' option in addition to the originally provided options. When LLM thought the question is unanswerable, it infers the additional information needed to answer, retrieves this information through the VQA model, and attempts to answer the question again. Section \ref{add-info inference} describes the method for inferring additional information, and Section \ref{loop-structure} explains the process for retrieving this information. Finally, to analyze the QA results from LLM, we constructed a sub-dataset by adding annotations to a portion of the videos in Social-IQ 2.0, the details of which are discussed in Section \ref{sub-dataset}.


\subsection{Inferring Additional Information with LLMs}
\label{add-info inference}
To enhance the reliability and interpretability of responses in QA tasks, we first allow LLM to determine whether it is possible to answer a question based solely on the input information related to the video such as retrived frames and automatic transcription. Specifically, we instruct LLM with prompts that outline how to respond in cases where it can answer the question and in cases where it cannot. When the question is judged answerable, LLM outputs the option it believes to be correct. If judged unanswerable, LLM outputs the additional information needed for answering, as well as the relevant time segments of the video. The additional information is selected from predefined options, including the ``scene context'', ``appearance of people'', ``facial expressions'', ``motion'', ``tone of voice'', and ``accurate dialogue''. The option ``accurate dialogue'' was included due to concerns that the accuracy might be insufficient based on our review of the actual transcript data. LLM infers the relevant time segments of the video based on the timestamps included in the dialogue information within the dataset, outputting the start and end times as integers.


\subsection{Retrieving Additional Information through the Loop Structure} 
\label{loop-structure}
Only when the additional information predicted in Section \ref{add-info inference} involves visual modalities, an image corresponding to the video reference time predicted by the LLM is retrieved from the dataset. Among the available options for additional information, the following four are classified as visual information: ``scene context'', ``appearance of people'', ``facial expressions'', and ``motion''. The timing for image retrieval is determined by rounding the average of the predicted start and end times to the nearest integer. 

The retrieved image is then input into the VQA model, along with prompts specifically defined for each modality, to generate detailed descriptions of the respective additional information. The visual information output by the VQA model is subsequently incorporated into the default prompt for the LLM (as explained in Section \ref{add-info inference}), and the model attempts to answer the question again. Notably, the ``unanswerable'' option is still retained in the second attempt, allowing for more reliable results compared to conventional methods.


\subsection{Construction of the Social-IQ Sub} 
\label{sub-dataset}
As mentioned earlier, the proposed method differs from the standard Social-IQ 2.0 approach in the following ways: (1) it introduces an ``unanswerable'' option, allowing for the retrieval of additional information when a question is deemed unanswerable; (2) it leverages tailored prompts to output the reasoning and justification behind answers, enabling an analysis of the model's response tendencies. To demonstrate the effectiveness of the proposed method and identify potential areas for improvement, we conducted an analysis that includes a comparison with human responses. To this end, we constructed a manually annotated dataset Social-IQ Sub.

{\bf Overview of the Annotation Process.} 
In the Social-IQ Sub, annotations were added to 200 videos randomly sampled from the Social-IQ 2.0 dataset: 174 from the training set and 26 from the validation set (This sampling ratio reflects the proportion of videos with full annotations in the downloaded Social-IQ 2.0 dataset). To address concerns that inaccurate transcripts might affect the precise evaluation of human response accuracy and the identification of necessary additional information (as discussed in \ref{add-info inference}), we generated more accurate transcripts using Whisper (large) \cite{whisper}, a state-of-the-art speech recognition model.

The annotation process was divided into two tasks: (1) answering questions based solely on the videos, and (2) answering questions based solely on the transcripts. Six annotators, three assigned to each task, completed the annotations over approximately one month. Quality control was performed in two stages: initially, a quality assurance reviewer checked the grammar and content, followed by a project manager who verified the number and format of the questions.

{\bf Annotation Items.} 
The primary aspects we aim to analyze with the newly constructed sub-dataset are as follows: (1) comparison of accuracy with human responses, (2) accuracy in determining answerability, (3) accuracy of the rationale behind responses, (4) accuracy of additional information retrieved when a question is deemed unanswerable, and (5) accuracy of the referenced video timestamps. To enable the analysis of these aspects, we defined specific annotation criteria for answers derived from both the video and the transcript.

For the video-only task, annotations were made for each question regarding answerability, the chosen option, the reasoning behind the answer, the information used as the basis for the answer (or additional information required if the question was deemed unanswerable), and the video time segment referenced for the answer. To facilitate analysis, predefined options were provided for the information used as the basis for the answer, with free-form responses allowed only when none of the predefined options applied. The predefined options included ``contents of conversation'', ``scene context'', ``appearance of people'', ``facial expression'', ``motion'', ``tone of voice'', and ``other information''.

For the transcript-only task, annotations covered answerability, the chosen option, the reasoning behind the answer, additional information required (only if the question was deemed unanswerable), and the video time segment referenced. The additional information options in this case included all items from the video-only task except for ``contents of conversation''.


\section{Experiments}
\label{sec:experiments}
\subsection{Experimental Setup} 
\label{sec:setup}
As previously mentioned, we used the Social-IQ 2.0 dataset \cite{siq2} for our experiments. Social-IQ 2.0 consists of 934 valid videos and 6,020 related questions. The number of videos is slightly reduced from the original dataset due to changes in video availability and region-based restrictions on YouTube. Each video is approximately 60 seconds long, with around six annotated questions per video. For each question, one correct answer and three incorrect answers are provided. The dataset also includes dialogue information (transcripts obtained via YouTube's automatic transcription feature), around 180 images extracted from each video at 3 frames per second, and an audio file. 

The LLMs used in this study were Llama 3 (8B-Instruct) \cite{llama3}, GPT-4 \cite{gpt4}, and GPT-4o \cite{gpt-4o}. To accommodate context window limitations, we used GPT-4 Turbo (hereafter referred to as GPT-4). BLIP-2 \cite{blip} was employed as the caption generation model, and GPT-4o served as the VQA model. 

In this experiment, we conducted QA on the validation data (120 videos, 807 questions) using the dialogue information (transcripts) and images. Since Llama 3 and GPT-4 cannot directly process images, we input the generated captions instead. The evaluation metrics were accuracy across all questions and accuracy across answerable questions.

\subsection{Accuracy Comparison Between Existing Methods and LLMs} 
We evaluated the social intelligence of LLMs by comparing their accuracy to existing methods. Specifically, we performed QA on 120 validation videos from the Social-IQ 2.0 dataset using Llama 3, GPT-4, and GPT-4o. To ensure a fair comparison with existing methods, we provided prompts that required the LLMs to produce an answer regardless of their certainty. Due to GPT-4o's limitation of processing only 10 input images, we used 10 evenly spaced frames extracted from each video. For GPT-4 and Llama 3, we used captions generated by BLIP-2 from the same 10 images. The results are presented in Table \ref{tab:1}.

For existing methods, we selected Just Ask Plus \cite{just-ask-plus}, which has the highest accuracy among non-learning methods, and the Multi-Modal Temporal Correlated Network with Emotional Social Cues (MMTC-ESC) \cite{multi}, which achieves the highest accuracy among learning methods. Additionally, we included the baseline framework (referred to as DeSIQ) developed in a study that focused on evaluating the Social-IQ dataset and creating a new dataset \cite{desiq}. DeSIQ processes features from each modality (question, correct answer, incorrect answers, transcript, or video) using a simple MLP. The accuracy of these existing methods is based on the values reported in their respective papers. Although each method utilized different modalities, to enable a more direct comparison, we excluded audio information and focused on visual information (images, captions, videos) and dialogue information (transcripts). Please note that the term ``video'' here refers to either the entire video or short clips segmented from the video.

\begin{table}[t]
  \caption{Comparison of QA accuracy between existing methods and LLMs.}
  \centering
  \begin{tabular}{ccc}
    \toprule
    Model & Modality & Accuracy \\ 
    \midrule
    Just Ask Plus \cite{just-ask-plus} & video-based & 53.4\% \\
    MMTC-ESC \cite{multi} & video-based & 74.35\% \\
    DeSIQ \cite{desiq} & video-based & 62.28\% \\
    \hline
    Llama 3 \cite{llama3}   & image-based & 52.66\% \\ 
    GPT-4 \cite{gpt4}       & image-based & 67.04\% \\ 
    GPT-4o \cite{gpt-4o}    & image-based & 75.22\% \\             
    \bottomrule
  \end{tabular}
  \label{tab:1}
\end{table}

\begin{table*}[t]
  \caption{Comparison of QA accuracy with and without loop structure.}
  \centering 
  \small 
  \scalebox{0.98}{
  \begin{tabular}{cccccc}
    \toprule
    Model (setting) & \# Correct  & \# Wrong & \# Unanswerable & Accuracy (overall) & Accuracy (answered questions) \\ 
    \midrule
    Llama 3 (original) \cite{llama3}  & 425 &  382   & - & 52.66\% &52.66\%\\ 
    Llama 3 (w/ ``unanswerable'')  \cite{llama3}       & 330   & 213   &264    & 40.89\%  & 60.77\% \\
    Llama 3 (LVD) \cite{llama3}       & 421   & 260   & 126   & 52.17\%  & 61.82\% \\       
    \hline
    GPT-4 (original) \cite{gpt4}      &  541  & 266 & - & 67.04\% &67.04\%\\ 
    GPT-4 (w/ ``unanswerable'') \cite{gpt4}  & 334 &  104  &  369  &  41.39\%  & 76.26\% \\ 
    GPT-4 (LVD)   \cite{gpt4}   &  372  &  124  &  311  & 46.09\%  & 75.00\% \\
    \hline
    GPT-4o (original) \cite{gpt-4o}   & 607 & 200  & - & 75.22\% &75.22\%\\ 
    GPT-4o (w/ ``unanswerable'') \cite{gpt-4o}         & 581   & 182   & 44    & 72.00\%  & 76.15\% \\ 
    GPT-4o (LVD) \cite{gpt-4o}        & 601   & 194   & 12     & 74.47\%  & 75.60\% \\
    \bottomrule
  \end{tabular}
  }
  \label{tab:2}
\end{table*}

\begin{figure*}[ht]
  \centering
  \includegraphics[width=\linewidth]{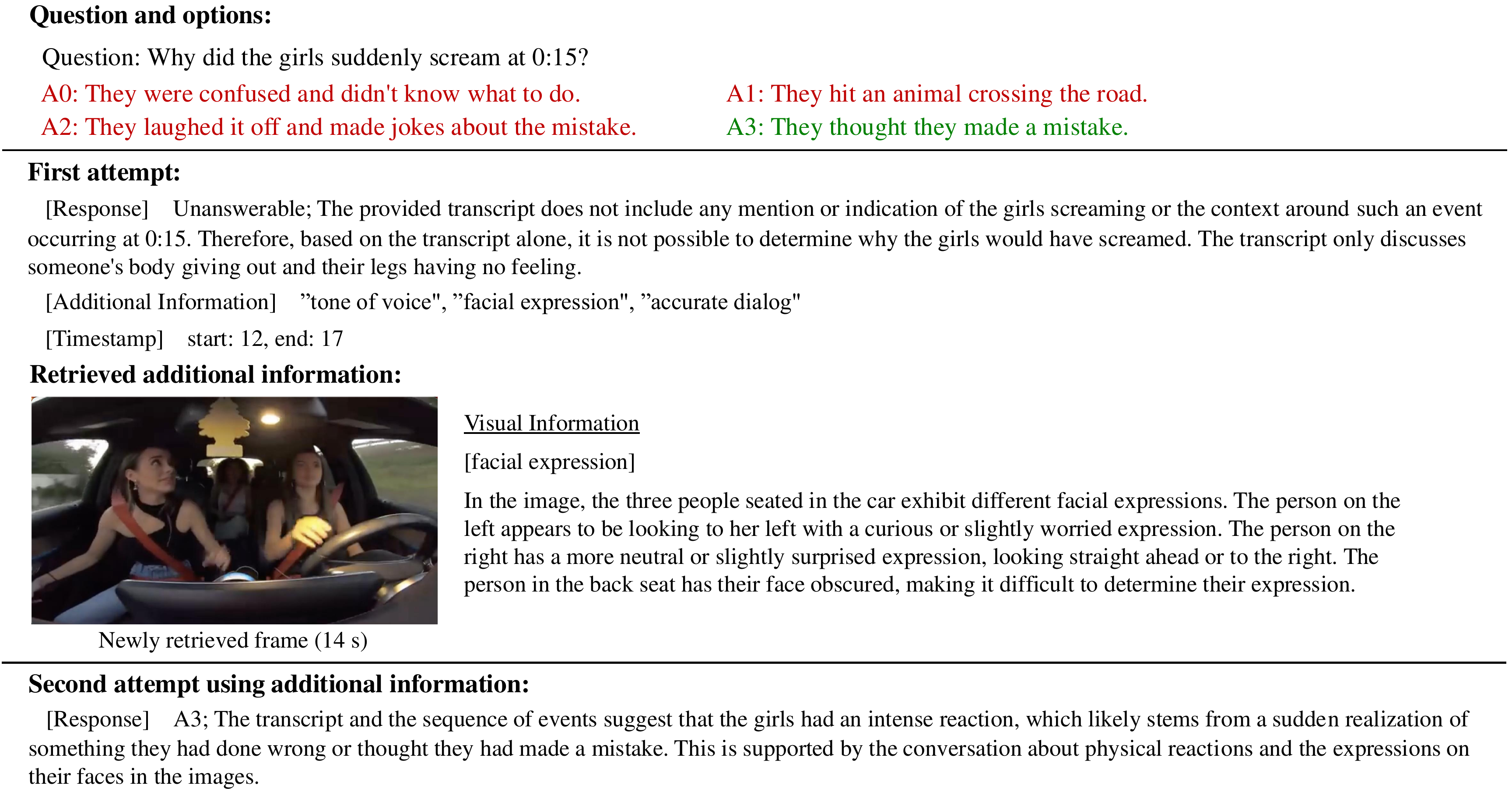}
  \caption{
  {\bf Result example.}
    In the ``Question and options:'' section, \textcolor{red}{red} options indicate incorrect answers, while \textcolor{OliveGreen}{green} options indicate correct ones. In this case, the first attempt resulted in ``Unanswerable'', but the correct answer was later produced using additional information from the VQA model. Under ``First attempt:'', the response, the additional information inferred by the LLM, and the corresponding video timestamps are recorded. Under ``Acquired additional information:'', the frame retrieved based on the predicted timestamps and the visual details obtained by the VQA from the frame is documented.
  }
  \label{fig:success}
\end{figure*}

As shown in Table \ref{tab:1}, the QA accuracy based on GPT-4o significantly outperformed the non-learning method (Just Ask Plus) and even surpassed the accuracy of the learning methods (MMTC-ESC, DeSIQ). Similarly, GPT-4 achieved far higher accuracy than the non-learning method, while Llama 3 produced results comparable to the non-learning method. However, it should be noted that Llama 3 could not be prompted to answer all questions, potentially lowering its accuracy score (94 out of 807 questions were left unanswered and counted as incorrect). These results suggest that LLMs have already acquired a certain degree of social intelligence. Additionally, GPT-4o's accuracy is 8.2\% higher than GPT-4, likely due to its multimodal capabilities and the information disparity between images and captions.

\subsection{Checking the Validity of the Loop Structure}

In this section, we assess the reliability of output answers by introducing an ``unanswerable'' option, a concept not previously explored in existing research. As shown in Table \ref{tab:2}, the inclusion of this option reduced accuracy by about 16\% for GPT-4 and 12\% for Llama 3, highlighting the impact of uncertain responses. In contrast, GPT-4o showed only a 3.2\% decrease, as it rarely chose ``unanswerable''. Additionally, Table 2 demonstrates that incorporating the loop structure improved QA accuracy by 2.5\% to 11.3\% compared to the non-loop scenario. This suggests that acquiring additional visual information inferred by the LLM helps answer complex questions related to social intelligence.
For qualitative results, actual examples of responses are provided in Figure \ref{fig:success}. As illustrated, when the LLM accurately infers the necessary information and the relevant time segments of the video, it can utilize the retrieved information to reach the correct answer.

\begin{table*}[t]
  \caption{
  Accuracy Comparison Between Human and the Proposed Method. (trans: transcript; video: full-length video; cap: captions of 10 frames; img: 10 images)}
  \label{tab:headings}
  \centering
  \small
  \begin{tabular}{ccccccc}
    \toprule
    Model             & Input    &\# Correct    &\# Wrong   & \# Unanswerable    & Accuracy (overall) & Accuracy (answered questions) \\ 
    \midrule
    Human             &trans     &488     &146     &663    &  37.63\%  & 76.97\% \\ 
    Human             &video     &997     &178     &122    &  76.87\%  & 84.85\% \\ 
    \hline
    Llama 3 (LVD) \cite{llama3}  & cap+trans &  669  &  360    &  268   & 51.58\%  & 65.01\% \\ 
    GPT-4 (LVD)  \cite{gpt4}     & cap+trans &  688  &  235    &  374   & 53.05\%  & 74.54\% \\
    GPT-4o (LVD)  \cite{gpt-4o}  & img+trans &  940  &  332     &  25   & 72.47\%  & 73.90\% \\
    \bottomrule
  \end{tabular}
  \label{tab:3}
\end{table*}


\subsection{Comparison of Human and LLM Responses}
\label{human-llm}
As an initial step in understanding the aspects humans focus on when interpreting interpersonal interactions and to enhance AI-based methods, we analyzed human responses and compared them with AI responses using the newly created Social-IQ Sub dataset.

The upper part of Table \ref{tab:3} shows the QA accuracy achieved by humans. To ensure a fair comparison between AI and human performance, we applied the proposed method to the Social-IQ Sub dataset and performed VideoQA, with the results displayed in the lower part of Table \ref{tab:3}. As shown, human accuracy was not 100\%, and in some ``unanswerable'' cases, the semantic distinctions between the options were unclear, or none of the options were appropriate. This highlights the inherent difficulty of designing suitable questions and choices for high-context benchmarks like Social-IQ. Therefore, it is essential for AI to reason like humans and help identify errors within the dataset.

Moreover, the fact that humans labeled more options as ``unanswerable'' than GPT-4o indicates that determining ``unanswerable'' with clear reasoning is challenging, underscoring a significant difference between LLMs and AI.

\begin{figure}[ht]
  \centering
  \includegraphics[width=\linewidth]{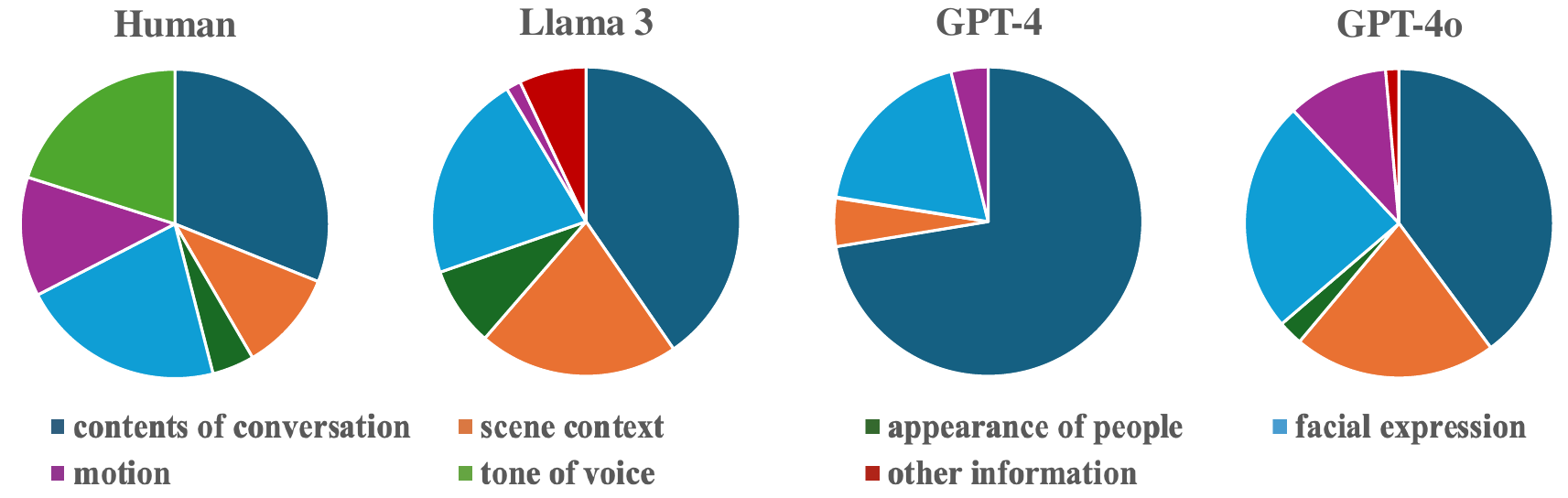}
  \caption{Comparison of rationale for answers between humans and LLMs.}  
  \label{fig:rationale}
\end{figure}

\begin{figure}[ht]
  \centering
  \includegraphics[width=\linewidth]{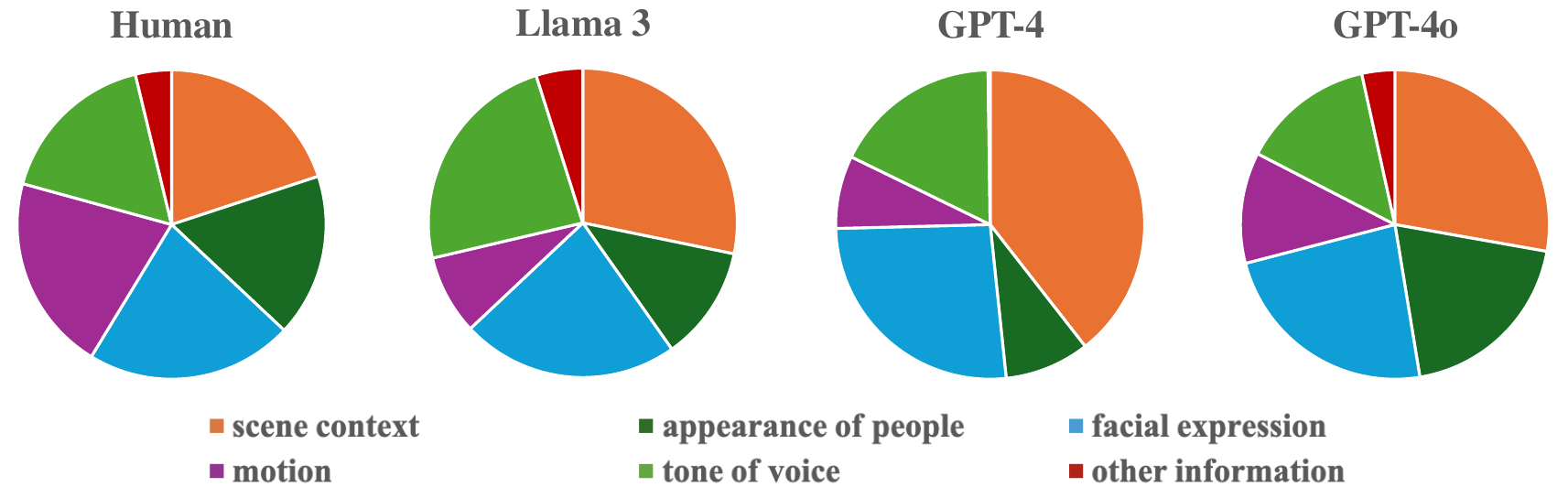}
  \caption{Comparison of additional information required by humans and LLMs.}  
  \label{fig:add-info}
\end{figure}

Next, we evaluated the accuracy of acquiring additional information within the loop structure and the correctness of the rationale provided for answers. Results related to the rationale for answers are shown in Figure \ref{fig:rationale}, while those related to additional information are presented in Figure \ref{fig:add-info}.

Figure \ref{fig:rationale} shows the frequency of information types used to answer questions based solely on video content. As indicated, Llama 3 and GPT-4o produced results relatively close to those of humans, suggesting a reasonable degree of accuracy in the rationale predictions by LLMs. However, these models tend to rely on easily available information, such as ``contents of conversation'' and ``scene context'', while providing less of the information like ``motion''. This suggests that the proposed method, which limits additional information to a single image, may not fully capture the information available in the video.

Figure \ref{fig:add-info} illustrates the frequency of additional information needed to answer questions deemed unanswerable using only transcripts. As shown, Llama 3 and GPT-4o produced results close to those of humans, indicating high accuracy in LLM predictions of additional information. However, Figures \ref{fig:rationale} and \ref{fig:add-info} show that GPT-4 deviates more from human judgment in both rationale and additional information compared to the other two LLMs. This may relate to the finding in Table 2 that, despite many unanswerable samples, GPT-4’s accuracy did not improve significantly with the loop structure.

\begin{table}[t]
  \caption{Video Reference Time Accuracy. }
  \label{tab:4}
  \centering 
  \scalebox{0.935}{
  \begin{tabular}{cccccc}
    \toprule
    Model & Input  & IoU  & \# Samples & \makecell{Avg\_length \\ (predicted)} & \makecell{Avg\_length \\ (annotated)} \\ 
    \midrule
    Llama 3 (LVD)     & trans   & 0.23  & 337 & 15.23 & 35.29 \\ 
    GPT-4 (LVD)       & trans   & 0.47  & 449 & 34.92 & 25.67 \\
    GPT-4o (LVD)      & trans   & 0.43   & 536 & 26.02 & 25.68 \\
    \hline
    Llama 3 (LVD)     & video   & 0.23  & 533 & 10.94 & 43.03\\ 
    GPT-4 (LVD)       & video   & 0.38  & 631 & 17.56 & 35.71\\
    GPT-4o (LVD)      & video   & 0.48  & 964 & 25.82 & 32.44\\
    \bottomrule
  \end{tabular}
  }
\end{table}

We calculated the IoU (Intersection over Union) between the reference times provided by LLMs and humans, with the results shown in Table \ref{tab:4}. Note that the prediction of reference timestamps is limited to samples that can be answered using either the transcript or the video. As indicated in Table \ref{tab:4}, the IoU values range from 0.2 to 0.5, suggesting there is room for improvement. Among the three LLMs, GPT-4o achieved the highest accuracy. In the annotated data, the average reference time for all samples was 11.50 seconds when using the transcript and 24.96 seconds when using the video. The average reference times shown in the table are longer than those for all samples, regardless of modality or LLM. This implies that while the proposed method is effective at capturing key information throughout the entire video, it has difficulty with questions requiring focus on specific timestamps.


\section{Conclusion} 

In this study, we evaluated the social intelligence of LLMs and improved accuracy on the Social-IQ 2.0 benchmark without fine-tuning by combining LLM capabilities with detailed visual information using a looped structure. Additionally, we created a new dataset by annotating human responses and their rationale (or additional information required) on a subset of Social-IQ 2.0 data, enabling a direct comparison between human and AI performance and contributing to enhancing AI's social intelligence capabilities. In the future, we plan to strengthen the framework by using Video LLMs as VQA models and incorporating audio modalities, based on insights gained from evaluating the newly created Social-IQ Sub dataset. 

\clearpage

\end{document}